\documentclass[10pt,twocolumn,letterpaper]{article}

\usepackage[pagenumbers]{cvpr} 

\usepackage{csquotes} 
\usepackage{xspace}
\usepackage{xcolor}
\usepackage{ulem}
\usepackage{amsmath}
\usepackage{algorithm}
\usepackage{algorithmic}
\usepackage{xcolor}
\usepackage{hyperref}

\definecolor{cvprpink}{rgb}{0.858, 0.188, 0.478}
\usepackage{booktabs}
\usepackage{pifont} 
\usepackage[table,xcdraw]{xcolor}
\usepackage{graphicx} 

\definecolor{subtlerow}{HTML}{F5F7FA}

\definecolor{lightblue}{HTML}{E6F0FF}
\hypersetup{
    colorlinks=true,
    linkcolor=cvprpink,
    urlcolor=cvprpink,
}

\newcommand{\paper}{EditYourself}
\newcommand{\projectpage}{edit-yourself.github.io}
\definecolor{myRed}{HTML}{C00000}

\newcommand{\stx}[2]{%
    \mbox{%
        {\color{myRed}\sout{#1}}
        \smash{\raisebox{-0.75ex}{\color{myRed}\fontsize{6pt}{6pt}\selectfont\textbf{#2}}}
    }%
}

\newcommand{\addx}[2]{%
    \mbox{%
        \textcolor{ForestGreen}{#1}
        \smash{\raisebox{-0.75ex}{\color{ForestGreen}\fontsize{6pt}{6pt}\selectfont\textbf{#2}}}
    }%
}

\newlength{\editcolwidth}
\definecolor{bestc}{rgb}{1.0, 0.92, 0.75}
\definecolor{secondc}{rgb}{0.93, 0.93, 0.93} 
\newcommand{\best}[1]{\cellcolor{bestc}\textbf{#1}}
\newcommand{\second}[1]{\cellcolor{secondc}\textbf{#1}}

\definecolor{cvprblue}{rgb}{0.21,0.49,0.74}
\hypersetup{breaklinks,colorlinks,allcolors=cvprblue}
\usepackage[hyperpageref]{backref}

\def\paperID{*****} 
\def\confName{CVPR}
\def\confYear{2026}

\title{\paper: Audio-Driven Generation and Manipulation of Talking Head Videos with Diffusion Transformers}

\author{
    John Flynn$^{1,*}$ \quad 
    Wolfgang Paier$^{1,*}$ \quad 
    Dimitar Dinev$^1$ \quad 
    Sam Nhut Nguyen$^1$ \\
    Hayk Poghosyan$^1$ \quad 
    Manuel Toribio$^1$ \quad 
    Sandipan Banerjee$^{2,}$\footnotemark[2] \quad 
    Guy Gafni$^{1,\ddagger}$ \vspace{0.2cm}\\
    $^1$ Pipio AI, $^2$ Amazon \\
    {\tt\small $^1$firstname.lastname@pipio.ai, $^2$sandgban@amazon.com} \\
    \small Project page: { \href{https://\projectpage}{\color{magenta}\texttt{\projectpage}}}
}

\begin{document}
\twocolumn[{%
\renewcommand\twocolumn[1][]{#1}%
\vspace{-3em}
\maketitle
\vspace{-2.4em}
\centering
\includegraphics[width=0.98\linewidth]{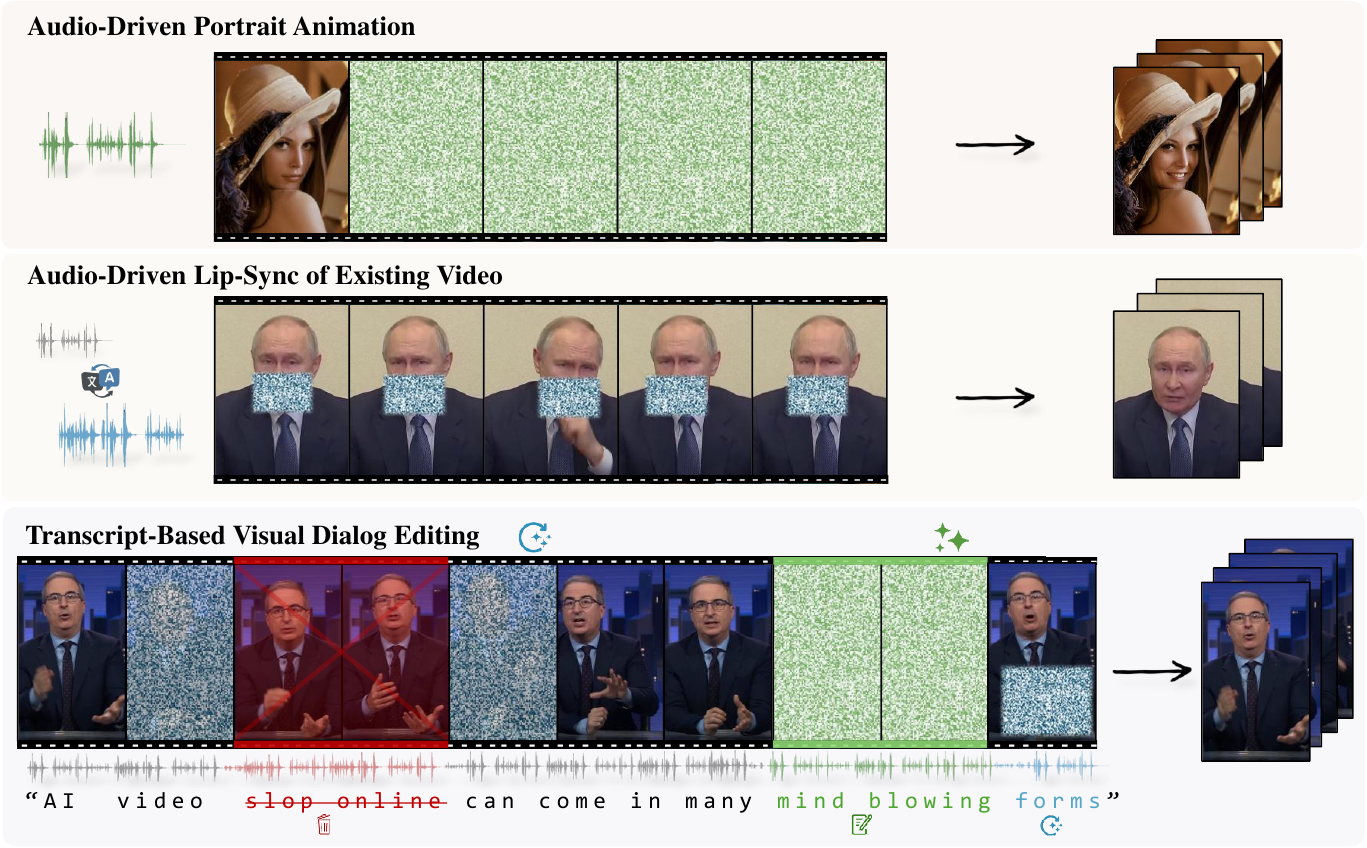}
\captionof{figure}{\paper \text{ }is a multipurpose lip-syncing video diffusion model designed for transcription-based dialog editing, capable of lip-syncing from a single frame or an existing video, and seamlessly editing the video to match the new script.} 
\label{fig:teaser}
\vspace{0.5cm} 
}]


\begin{abstract}

\vspace{-1mm}
Current generative video models excel at producing novel content from
text and image prompts, but leave a critical gap in editing existing pre-recorded videos, where minor alterations to the spoken script require preserving motion, temporal coherence, speaker identity, and accurate lip synchronization. We introduce \paper, a DiT-based framework for audio-driven video-to-video (V2V) editing that enables transcript-based modification of talking head videos, including the seamless addition, removal, and retiming of visually spoken content. Building on a general-purpose video diffusion model, \paper\text{ }augments its V2V capabilities with audio conditioning and region-aware, edit-focused training extensions. This enables precise lip synchronization and temporally coherent restructuring of existing performances via spatiotemporal inpainting, including the synthesis of realistic human motion in newly added segments, while maintaining visual fidelity and identity consistency over long durations. This work represents a foundational step toward generative video models as practical tools for professional video post-production.
\insert\footins{\noindent\footnotesize $^*$Authors contributed equally.}
\insert\footins{\noindent\footnotesize $^\dagger$Work done while at Pipio AI.}
\insert\footins{\noindent\footnotesize $^\ddagger$Project Lead.}
\end{abstract}    
\section{Introduction}
\label{sec:intro}

A growing share of modern video content is human-centric, including movies, online courses, corporate communications, interviews, and short-form social media uploads. In these videos, creators often need to revise the spoken content post-recording to fix fumbled lines, update facts, remove filler words, tighten interviews or localize content across languages. Traditional non-linear video editing tools, however, provide limited support for such edits, as operations like inserting, removing or retiming speech typically introduce visible jump cuts or unnatural motion. Moreover, selectively re-rendering only parts of a human performance requires extensive manual intervention within existing post-production workflows~\cite{Fried:text_based_editing_talking_heads,ChunkyEdit,VideoDiff}. Recent advances in video diffusion models suggest a promising alternative. These models can synthesize high-quality, temporally coherent human videos from text, images or audio, demonstrating an ability to model complex appearance, motion, and facial dynamics~\cite{imagen_video,ltx}. This capability makes generative models a good candidate for not only content creation, but as editing engines that can repair, extend or reshape existing videos in a content-aware manner~\cite{dreamix}.

However, research on editing existing human-centric content remains far less mature than work on end-to-end generation. Most current approaches focus on Image-to-Video (I2V) generation from a single portrait image~\cite{hallo3,sonic,echomimic,memo}. While impressive in their realism, these methods frequently suffer from identity drift over time and incorrectly reproduce a subject's likeness. A single image cannot capture the full range of facial details and speaking style present in a real performance. As a result, generated videos often hallucinate details such as teeth, wrinkles, facial hair or gestures, producing outputs that feel incorrect, especially when users are generating videos of themselves. On the other hand, V2V lip-sync models~\cite{wav2lip,latentsync,musetalk,infinitetalk} adhere closely to an input video to preserve visual fidelity and identity, but offer limited flexibility for editing. By operating under a fixed temporal structure that preserves the original frame count and timing, these methods make it difficult to insert or remove speech segments while maintaining temporal continuity. Consequently, existing V2V and I2V methods do not adequately support precise edits required for real-world post-production workflows.

Our work tackles this fundamental problem of temporal manipulation of existing talking-head videos, which we refer to as \textit{visual dialog editing}: V2V editing driven by changes to the spoken dialog~\cite{text_based_editing}. This setting goes beyond simple lip synchronization to completely new audio, and supports core post-production operations such as inserting, removing and retiming video segments while preserving visual continuity. Editing videos directly through their textual transcript provides an intuitive and expressive interface for creators, enabling precise word-level modifications such as filler-word removal and post-shoot script revisions. More broadly, this transcript-centric workflow shifts video production from a ``script-perfect-before-shooting” paradigm toward a ``shoot once, refine later'' model, enabling rapid updates, personalized variants and integration with higher-level control systems such as LLM-based AI agents for automated video editing~\cite{ChunkyEdit,podreels,VideoDiff}.

In this work, we address this gap by re-framing talking-head video synthesis as a problem of visual dialog editing. We introduce \textit{\paper}, a diffusion-based framework designed specifically for transcript-driven editing of talking head videos. By adapting a pre-trained general-purpose video diffusion model into a flexible, audio-driven V2V editor, our approach enables precise modification of existing videos, including addition, removal, and retiming of spoken segments, while maintaining accurate lip synchronization, visual identity, and temporal coherence over long videos.

In summary, our work makes the following contributions:
\begin{itemize}
    \item \textbf{Lip-sync on a pretrained video diffusion model:} We introduce a two-stage training scheme that enables inference on speech audio across varying text, image, and video inputs, while maintaining accurate lip synchronization, together with a windowed audio conditioning strategy for precise speech-video alignment that does not require audio feature downsampling and remains robust across varying video frame rates.
    \item \textbf{Latent-space visual dialog editing:}
    We formulate transcript-driven video editing directly in latent space, supporting seamless addition, removal, and retiming of spoken segments.
    \item \textbf{Identity-preserving long video generation:}
    We introduce a reference-based identity conditioning mechanism, \textit{Forward–Backward RoPE Conditioning}, together with TeaCache-aware inference, to stabilize appearance and temporal coherence over long videos.
\end{itemize}

Evaluations against recent I2V and V2V lip-sync benchmarks demonstrate that our method achieves SOTA visual quality and synchronization accuracy. In addition to offering competitive performance, our approach represents a foundational step toward utilizing video diffusion models as capable tools for editing human-centric video content.

\section{Related Works}
\label{sec:related_works}
With the advent of diffusion models~\cite{diffusionmodelprinciples, lipman2023flowmatchinggenerativemodeling}, the field of video generation~\cite{vidgensurvey} has proliferated in recent years. Coupled with powerful 3D VAEs~\cite{vae_paper}, these models have the capability of reconstructing the details and dynamics of an entire frame (instead of a small crop), opening the door to generating novel frames that are coherent with the rest of the video. There are several possible input modalities, which can be combined together, that define the task of the model. The common modalities are: \textbf{(i)} {\textit{Text-to-Video} (T2V)} synthesize the video from a textual input~\cite{nuwa_eccv22, cogvideo, imagen_video}, \textbf{(ii)} {\textit{Image-to-Video} (I2V)} animate a single image into a video~\cite{lumiere}, \textbf{(iii)} {\textit{First-Last-frame-to-Video} (FL2V)} guide video generation between the given first and last frames~\cite{deep_video_prior, veo_paper}, and \textbf{(iv)} {\textit{Video-to-Video} (V2V)} edit or transform video content while maintaining temporal consistency and structure.~\cite{infinitetalk, flowvid}. The latest video generation works~\cite{wans2v, sora_tech_report, ltx} focus on synthesizing clips that adhere to provided prompts by leveraging diffusion transformer blocks (DiTs)~\cite{dit_iccv} as the main computational units in their models. A newer version of this, multi-modal diffusion transformer blocks (MM-DiTs)~\cite{sd3_mmdit} allow multiple input modalities to be represented in a common token space, facilitating joint attention across them.

\subsection{Audio-Driven Talking Head Generation}
\paragraph{Early Methods.}Audio-driven facial animation, in particular lip-syncing, has been an active research topic with a variety of methods explored. GAN-based methods~\cite{lipgan,lip_movement_gen_eccv18,stylesync,stylegan2,dimitar_cvpr_23} achieved early success with appropriate audio representations, such as Wav2Lip and Wav2Vec~\cite{wav2lip, wav2vec} for conditioning. These methods can indeed lip-sync a video, however are unable to make larger changes like head motion. Adding 3D Morphable Models~\cite{blanz2003face} from traditional graphics as an intermediate representation allows enhanced control over the subject in the video. Lip-sync and head control can be added using the parameterizations (\eg blendshapes) of the models~\cite{voca,meshtalk,codetalker}. Volumetric rendering techniques from graphics~\cite{lombardi2019neural,nerf} have found success in representing human avatars~\cite{guy_cvpr_21,adnerf,dfa_nerf,gene_face}. More recently, 3D Gaussian Splatting~\cite{3dgs} techniques have also been used to successfully lip-sync videos~\cite{gaussiantalker, talkinggaussian}.
\paragraph{Diffusion based Methods.}In the last couple of years, latent diffusion models~\cite{ldm} have become the backbone of choice for generating talking head videos from a single source image or video. The earlier set of these models typically use a pre-trained 2D/3D VAE~\cite{vae_paper} to encode the source and a trainable UNet-style module~\cite{unet} for denoising. A trainable copy of the UNet acts as a reference net to inject control signals into the denoiser's feature space. These signals can be audio representations~\cite{wav2lip,wav2vec}, emotion embeddings~\cite{memo}, face and body keypoints~\cite{echomimic, animateanyone,keysync} or identity information~\cite{hunyuanportrait,keyface}. However, recent models replace the denoising UNet with a diffusion transformer (DiT)~\cite{dit_iccv} for improved scalability and global context handling~\cite{mocha,hallo3,fantasytalking,stablevideoinfinity,omnisync,wans2v,multitalk}, and focus on multi-stage training~\cite{omnihuman_1, omnihuman_15} where the model is incrementally trained on a higher data dimensionality of the source (\eg audio, then image, then video). The final model can then be controlled by only audio~\cite{sonic,omniavatar} or combining it with blendshapes~\cite{omnitalker}, pose information~\cite{humandit,playmate,omnisync}, external embeddings~\cite{hunyuanvideoavatar,resyncer,fantasytalking} that is injected into the latent model via cross attention or a multi-modal block~\cite{sd3_mmdit}. The majority of these models~\cite{stableavatar,rap,wans2v,mocha} use a flow-matching objective~\cite{lipman2023flowmatchinggenerativemodeling} rather than denoising diffusion due to its faster sampling and straightforward noise to data path.

\subsection{Video Manipulation}
\paragraph{Video-to-Video Editing.}As a consequence of the above line of research, V2V editing and manipulation applications have gained considerable popularity of late. Some of these models, like Runway's Gen-3 Alpha/Gen-4.5\footnote{\url{https://runwayml.com/research/introducing-runway-gen-4.5}}, perform style transfer~\cite{Johnson2016Perceptual} on existing videos while others (\eg~\cite{wan22animate},~\cite{unianimate},~\cite{animateanyone},~\cite{alignhuman}) focus on direct motion transfer from conditioning signals. For explicit content insertion/removal from video frames, inpainting models leveraging optical flow have been explored~\cite{fgt,propainter}, with recent works using diffusion~\cite{lumiere,avid} and flow matching~\cite{ditpainter,vace}.

The video editing problem can be formulated as a collection of image editing steps (or ``slices''~\cite{slicedit}) directly using a pretrained T2I model. Finetuning the model~\cite{factorized_diffusion_distillation, dreamix, ccedit} enables text-based editing, while enhancements such as feature banks and optical flow~\cite{avid,stream_v2v,flowv2v,video_handles} can improve restyling quality and object removal. Latent diffusion models~\cite{ldm} are suitable for surgically editing specific regions of the video, as there is a clear mapping between the space and time coordinates of any given video pixel to the latent tokens it generates~\cite{vfrtok}. Keyframes, a concept from traditional video editing, can also be used to loosen the one-to-one correspondence between the input and output video~\cite{infinitetalk, keyface, keysync} and produce videos that match the subject, but could have different head or hand movements.

\paragraph{Transcript-Based Editing.}A problem closely related to our work is how to handle changes in the spoken script without requiring re-shooting of the whole segment. Early methods~\cite{Fried:text_based_editing_talking_heads} presented a dynamic programming-based synthesis strategy to assemble new speech videos combining visemes, 3DMM-based blending, and a recurrent video generation network, while~\cite{yao2021talkinghead} used a fast phoneme search and neural re-targeting to transfer mouth movements from the source to a target. The talking-head editing process can also be broken down into audio-to-dense-landmark motion and motion-to-video stages~\cite{cascaded_conditioning, yang2023contextawaretalkingheadvideoediting}. Although these methods enable transcript-based editing, they are limiting in that they either require subject-specific data or they struggle to generalize to diverse videos.

\subsection{Identity-Preserving Long Video Generation}
Generating longer videos with diffusion models remains a technical challenge. The spatio-temporal dimensions of the output video are determined by those of the noise tensor (\ie the sequence length of the noise tokens), practically limited by GPU memory. Naive auto-regressive techniques experience drastic reductions in video quality and identity preservation~\cite{self_forcing}. Recent techniques mitigate video quality degradation~\cite{self_forcing, stablevideoinfinity, exposure_bias, framepack}, but these are not sufficient to prevent identity drift in human faces (\ie loss of facial details, over-smoothing, over-saturation of the skin, changes in facial hair). A simple approach is to leverage a reference subject image encoding, as done in~\cite{hunyuanvideoavatar, wan22animate, lookahead, hallo3, jiang2024loopy, omnihuman_1, zhou2024storydiffusion}. As the reference image can contain background information not related to the subject's identity, embeddings from CLIP~\cite{CLIP} or the face-specific ArcFace model~\cite{arcface} can also be used~\cite{consis_id, hunyuanportrait, gan2025omniavatar, hong2025audio, magic_mirror}. These features can be integrated into the DiT via cross-modal adapters.

\section{Method}
\label{sec:method}

We base our model on LTX-Video~\cite{ltx}, a general-purpose video diffusion model that supports text, image, and video-conditioned generation (T2V, I2V, and V2V), which we introduce in Section~\ref{subsec:flowmatching}. Building on this backbone, we introduce a set of extensions that specialize the model for audio-driven and transcript-based video editing. Specifically, we include (i) cross-modal audio conditioning and a V2V lip-sync training strategy (\ref{subsec:lipsync}), (ii) a latent-space formulation of visual dialog editing that supports transcript-driven addition, removal, and retiming of speech (\ref{subsec:dialogediting}), (iii) a caching-aware long-inference strategy for temporally consistent generation over long durations (\ref{subsec:long_inference}), and (iv) reference-based identity conditioning with a novel Forward-Backward Rotary Positional Embedding (RoPE)~\cite{su2023roformerenhancedtransformerrotary} mechanism to stabilize appearance across both edited and fully synthesized segments (\ref{subsec:id_cond}).

\begin{figure*}[]
    \centering
    \includegraphics[width=0.92\linewidth]{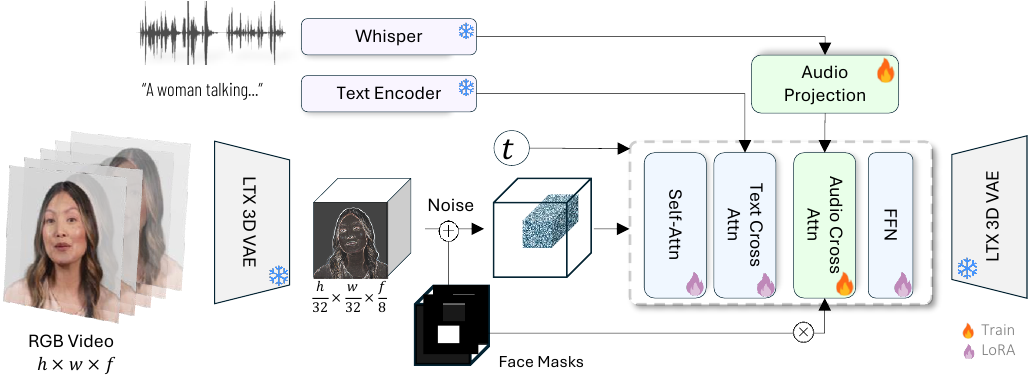}
    \caption{Our proposed pipeline. A global audio projection layer and audio cross-attention layers are added to the network's architecture. For V2V lip syncing, we apply noise to tokens corresponding to the mouth area and task the model with spatio-temporally inpainting them. }
    \label{fig:method}
\end{figure*}

\subsection{Preliminaries} 
\label{subsec:flowmatching}
\paragraph{Baseline Network.}We use the LTX-0.9.7~\cite{ltx} DiT and the associated Video-VAE as our baseline, which follows the common 3D causal VAE and flow-matching DiT pattern for video generation, including 3D RoPE~\cite{su2023roformerenhancedtransformerrotary} for spatio-temporal positions and adaptive normalization for timestep conditioning. With 14B parameters, the DiT model operates in a highly compressed latent space using a separately pre-trained Video-VAE. The VAE encoder's rather aggressive compression rate ($32 \times 32 \times 8$) results in a significantly lower token count, aimed at increasing performance towards interactive applications. Videos are generated in a two-pass fashion: (i) denoising is first performed on a coarser, lower-resolution representation of the video, (ii) followed by a learned upsampling of the latents and a second, higher-resolution denoising pass. Crucially, LTX-Video was pre-trained with a flexible multi-task objective spanning T2V, I2V, keyframe generation, and various forms of spatial and temporal inpainting. This is achieved by masking tokens and assigning them distinct conditioning timesteps, regardless of the global diffusion step.

\paragraph{Flow Matching Training Objective.}We adopt the Flow Matching~\cite{lipman2023flowmatchinggenerativemodeling} paradigm, following LTX-Video~\cite{ltx}. Formally, video samples are encoded into a latent representation $\mathbf{x}_0 \sim p_{\text{data}}$ using the LTX-Video Video-VAE. We define a linear probability path to interpolate between latent video representation $\mathbf{x}_0$ and a noise distribution $\mathbf{x}_1 \sim \mathcal{N}(0, \mathbf{I})$ via the displacement flow $\mathbf{x}_t = (1 - t)\mathbf{x}_0 + t \mathbf{x}_1$ with a continuous time step $t \in [0,1]$. The DiT, $v_\theta$, is trained to predict the velocity field that transforms noise back into data by minimizing our base training objective
\begin{equation}
\mathcal{L}_{\text{FM}} = \mathbb{E}_{t, \mathbf{x}_1, \mathbf{x}_0, \mathbf{c}} [\| v_\theta(\mathbf{x}_t, t, \mathbf{c}) - (\mathbf{x}_1 - \mathbf{x}_0) \|^2_2],
\end{equation}
where $\mathbf{c}$ denotes the available input conditions (text prompt, in the LTX-Video base model). In the subsequent subsections, we modify this objective to include audio and identity conditioning. Please see Equation~\eqref{eq:fulltrainingobjective} for the expanded training loss formulation.

At inference, new videos can be generated by solving the probability flow ODE,
$\frac{d\mathbf{x}_t}{dt} = v_\theta(\mathbf{x}_t, t, \mathbf{c})$, which requires integrating the velocity field from $t=1$ to $t=0$: 
\begin{equation}
\mathbf{x}_0 = \mathbf{x}_1 - \int_{0}^{1} v_\theta(\mathbf{x}_t, t, \mathbf{c}) dt
\end{equation}

In practice, this integration is discretized using a first-order Euler solver over $40$ steps following the update rule $\mathbf{x}_{t_{i-1}} = \mathbf{x}_{t_i} - \Delta t \cdot v_\theta(\mathbf{x}_{t_i}, t_i, \mathbf{c})$.
 For further details, please refer to the original LTX paper~\cite{ltx} and repository~\cite{ltx_github}. 

\subsection{Cross-Modal Audio \& Video Conditioning}
\label{subsec:lipsync}
\paragraph{Audio Conditioning Strategy.} Inspired by DiT-based portrait animation methods 
~\cite{sonic,omnisync,omnihuman_1,mocha,fantasytalking,rap}, we extend a pre-trained video diffusion model with an audio conditioning modality by introducing additional cross-attention layers into the transformer blocks. Specifically, we insert one such layer into each DiT block, positioned between the text cross-attention and the FFN. As keys and values, we use pre-extracted \texttt{Whisper-small}~\cite{whisper} features $\mathbf{c}_\mathrm{audio} \in \mathbb{R}^{L \times B \times C}$ with $L$ the sequence length, $B$ the number of encoder block outputs and $C$ the channel dimension. The proposed conditioning mechanism however is agnostic to the choice of audio representation. The audio features are processed by a learned projection and pooling module (Audio Projection) to produce lip-sync embeddings at the latent video frame rate. These embeddings are then shared across all DiT blocks, allowing audio information to modulate video features at every layer while preserving the pretrained DiT’s token structure.

To minimize disruption to the pretrained DiT at the start of training, we initialize the Audio Projection module's convolution layers as average pooling operators and set the audio cross-attention output projections to zero. During training, we randomly drop audio conditioning with probability $\overline{p}_\mathrm{audio}$ by detaching these layers. Overall, the Audio Projection module and associated cross-attention layers introduce approximately 2B additional learnable parameters. The resulting architecture is illustrated in Figure~\ref{fig:method}.

To restrict attention to temporally local audio context, we associate each video frame index $i$ with a window of $W$ audio features $\tilde{\mathbf{c}}_\mathrm{audio}^i \in \mathbb{R}^{W \times B \times C}$. Because the sampling rate of audio features (\eg Whisper embeddings) $f_a$ typically differs from the video frame rate $f_v$, naively selecting the nearest audio features can introduce sub-frame audio-video misalignment. Prior approaches often address this mismatch by interpolating audio features to $f_v$. However, this strategy is fragile for two reasons: (1) downsampling can discard high-frequency information present in modern speech embeddings, and (2) fixed-size windows correspond to different temporal durations across videos with varying frame rates.

To address these issues, we sample audio features on a phase-shifted grid that preserves the original audio feature rate $f_a$ while aligning audio windows to video frames. Specifically, for each video frame index $i$, we extract a center-aligned window of $W$ audio features from $\mathbf{c}_\mathrm{audio}$ at fractional audio indices $u_n$ using linear interpolation, yielding $\tilde{\mathbf{c}}_\mathrm{audio}^i[n]$, where $n$ denotes the index within the window.
\begin{align}
u_n &= i\,\tfrac{f_a}{f_v} + \left(n-\tfrac{W-1}{2}\right), n=0,\ldots,W-1,\nonumber\\
k_n &= \lfloor u_n \rfloor,\nonumber\\
\alpha_n &= u_n-k_n,\nonumber\\
\tilde{\mathbf c}^{\,i}_{\mathrm{audio}}[n]
&= (1-\alpha_n)\,\mathbf c_{\mathrm{audio}}[k_n] + \alpha_n\,\mathbf c_{\mathrm{audio}}[k_n+1] \label{eq:audio_resample}
\end{align}
This design decouples audio temporal resolution from video frame rate, ensuring consistent window semantics across videos with arbitrary frame rates.

To encode relative position within the audio window, we introduce a learned, fixed-size positional embedding tensor $\mathbf{P} \in \mathbb{R}^{W \times B \times C}$, where each slice $\mathbf{P}[n]$ corresponds to a window index.
\begin{align}
\tilde{\mathbf c}^{\,i}_{\mathrm{audio+pos}}[n]
&= \tilde{\mathbf c}^{\,i}_{\mathrm{audio}}[n] + \mathbf{P}[n]
\end{align}

\begin{figure}[h]
    \centering
    \includegraphics[width=\linewidth]{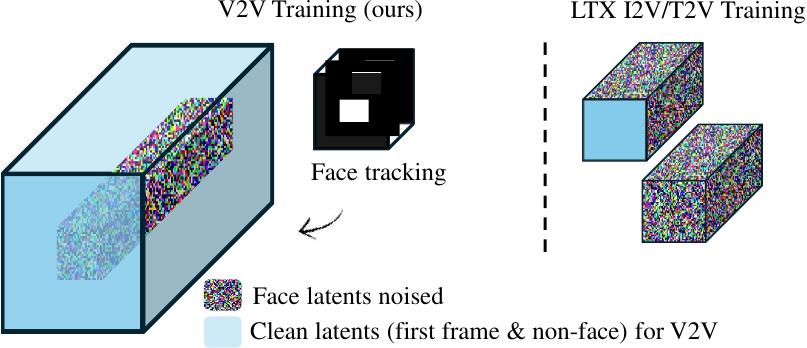}
    \caption{In order to train the audio attention layers, we fully noise the tokens corresponding to mouth region throughout the training sample. We retain clean latents of the first frame, similar to image-to-video training in LTX. The model learns to in-paint the mouth through time and space using the audio, and the initial mouth shape as conditions. }
    \label{fig:v2v}
\end{figure}

\paragraph{V2V Lip-Sync.}
LTX-Video~\cite{ltx} is a general-purpose video diffusion model supporting text, image, and video-conditioned generation. In its standard I2V usage, the model conditions on clean latents for an initial frame and noisy latents for subsequent frames, with self-attention propagating information to guide temporal generation. We build on this mechanism to specialize the model for audio-driven V2V lip synchronization by selectively regenerating the mouth region in talking-head videos. For each source video, we detect lower-face bounding boxes using MediaPipe~\cite{mediapipe48292} and compute an enclosing box over groups of eight consecutive frames, yielding a binary mask $\mathbf{M}$ per latent frame. During training, noise $\epsilon$ is applied at timestep $t$ only within the masked region $\mathbf{M}$, and the model is trained to inpaint the corresponding tokens over space and time while preserving unmasked content. See \cref{fig:v2v}.
\begin{equation}
\mathbf{x}_t = \mathbf{M} \odot [(1 - t)\mathbf{x}_0 + t \epsilon] + (1 - \mathbf{M}) \odot \mathbf{x}_0
\end{equation}

We also restrict the audio cross-attention layers to only update tokens belonging to the face region by multiplying the cross-attention output with the face mask:
\begin{equation}
\mathbf{z}_{\text{out}}
=
\mathbf{z}_{\text{in}}
+
\mathbf{M} \odot
\mathrm{AudioAttn}\!\left(\mathbf{z}_{\text{in}}, \mathbf{c}_a\right)
\end{equation}
where $\mathbf{z}$ denote the hidden latents in the DiT. 

We further apply random conditioning dropout to ensure that the model remains robust to missing spatial and temporal inputs, and can operate under different combinations of conditioning signals. Following the original training procedure for LTX-Video, we randomly drop first-frame conditioning with probability $\overline{p}_\mathrm{ff}$, reducing the objective to text-to-video generation. Similarly, we randomly drop video-to-video conditioning with probability $\overline{p}_\mathrm{v2v}$ to preserve the model's image-to-video generation capability. When video-to-video conditioning is absent, we set the spatial mask to $\mathbf{M}=1$, allowing audio cross-attention to update all tokens, enabling unconstrained latent generation. \cref{tab:training_parameters} reports the values we chose for  $p_\mathrm{ff}$, $p_\mathrm{v2v}$ and $p_\mathrm{audio}$.

\begin{figure}[h]
    \centering
    \includegraphics[width=0.75\linewidth]{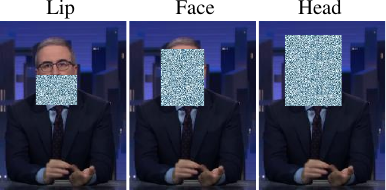}
    \caption{V2V Inference Modes. Adjusting the mask $\mathbf{M}$ in inference enables different synchronization levels: \textit{Lip} for mouth-only sync, \textit{Face} for expressions, and \textit{Head} to synthesize new head dynamics matching the audio prosody.}
    \label{fig:render_modes}
\end{figure}

Our masked training strategy affords substantial flexibility at V2V inference time. By selectively scaling and positioning the mask $\mathbf{M}$ (see \cref{fig:render_modes}), the model can be configured to synchronize only the lips, the face, or the entire head. In the \textit{Head} mask mode, the model leverages its generative prior over head motion learned during T2V and I2V lip-sync training, selectively re-synthesizing head dynamics to match the timing and prosody of the new speech. In contrast, the \textit{Face} and \textit{Mouth} modes progressively constrain generation to smaller spatial regions, producing new content within the masked area while increasingly adhering to the original video outside the mask.

\vfill\null
While noise levels are sampled individually for each token, we must bias the sampling of noise timesteps towards the high-noise region. If noise levels in the mouth region are too low during training, the pre-trained DiT’s inherent ability to denoise video tokens will allow it to infer the mouth correctly without relying on the audio signal, leading to a trivialization of the cross-attention layers in training, causing a collapse of the lip-sync task when inferencing on novel audio. This noise bias aligns with the observation of previous works that crucial lip-sync details are determined primarily in earlier stages of denoising ~\cite{alignhuman, omnisync, wan22animate}. We use a shifted log-normal distribution with shift $\mu = 2.05$ which places $90\%$ of timesteps in the range $[0.60, 0.98]$. The tendency of the lip shapes and mouth movements to be determined during the early noise stages is further validated during inference.

\subsection{Visual Dialog Editing}
\label{subsec:dialogediting}

Modification of the transcript in text domain admits changes to both audio and video domains. We address the visual synthesis challenge and use commercially available solutions for zero-shot voice cloning~\cite{elevenlabs2023,deepdub2023}. We target the generation of new frames, re-generation of existing frames, eliminating discontinuity artifacts across edit boundaries (jump-cuts), all while ensuring accurate lip synchronization to the target audio. Specifically, we define the following operations for a video segment:
\begin{enumerate}
    \item \texttt{Addition}: Insertion of new content at arbitrary timestamps, seamlessly adhering to surrounding boundary frames (when present).
    \item \texttt{Removal}: Deletion of existing content while smoothing the resulting temporal discontinuity to avoid visible jump cuts.
    \item \texttt{Re-render}: Selective inpainting of video content over specified spatial and temporal regions (\eg correcting an awkward facial expression or replacing a hand gesture).
    \item \texttt{Retime}: Altering the total duration of a video segment to match changes in script duration (\eg for language localization), implemented via evenly-distributed additions/removals.
\end{enumerate}

Unlike Addition and Removal, which are localized operations, Retime applies distributed temporal adjustments across the entire segment. This distinction is particularly important for dubbing, since translated speech often involves changes in word order, density, and duration; strict correspondence with the video timeline is lost, requiring adjustments to the overall duration of the segment rather than at specific words. We motivate the operations with the following examples in ~\cref{script_driven_manipulation_1,script_driven_manipulation_2}.

\begin{figure}[h]
    \small
    \centering
    \begin{tabular}{rp{\editcolwidth}}
        \textbf{Original:} & 
        \parbox[t]{\editcolwidth}{\enquote{This feature rocks and we will most likely launch it.}} \\\\
        \hline
        \textbf{Revised:} & 
        \parbox[t]{\editcolwidth}{\enquote{This 
        \addx{awesome new}{(+0.9s)}
        feature rocks and we will 
        \stx{\textcolor[HTML]{C00000}{most likely}}{(-0.5s)} 
        launch it \addx{next week}{(+0.6s).}} 
        }
    \end{tabular}
    \caption{Example of a script-driven temporal edit, illustrating the complexity of $\text{V2V}$ operations. New content is highlighted in green, and a redaction is shown with a red strike-through, accompanied by the required duration change for each operation. Two addition operations and a removal operation are needed to account for these edits.} 
    \label{script_driven_manipulation_1}
\end{figure}

\begin{figure}[h]
    \small
    \centering
    \begin{tabular}{rp{\editcolwidth}}
        \textbf{EN:} & 
        \parbox[t]{\editcolwidth}{\enquote{It's stated in our terms and conditions.}} \\
        \hline
        \textbf{DE:} & 
        \parbox[t]{\editcolwidth}{\enquote{Das ist in unseren Allgemeinen Geschäftsbedingungen festgelegt. \textcolor{ForestGreen}{ (+1.1s)}}} 
    \end{tabular}
    \caption{Example of a $\texttt{Retime}$ operation needed for language localization/dubbing. The English phrase expands significantly when translated into German, requiring the model to expand the duration of the entire segment by approximately $\texttt{+1.1s}$ to maintain natural speech.}
    \label{script_driven_manipulation_2}

\end{figure}

We obtain word-level timestamps using an automated transcription tool~\cite{deepgram2025,aws2025transcribe}. The user-edited transcript is then diff-checked against the original to identify changed spans and map them to their corresponding audio timestamps. This results in a finalized set of Addition, Removal, and Retime operations.

 We leverage LTX-Video's architectural flexibility and formulate visual dialog editing as a specialized inpainting task. To realize these edits, we modify the latent video frame sequence directly along the spatial and temporal axes. Frame addition is implemented by inserting fully noised latent frames at the corresponding locations, while frame removal deletes existing latent frames from the sequence. Exploiting the causality of the VAE encoder, we define a mapping between each latent frame at index $n$ and its corresponding range of input video frames indexed from $8(n-1)+1$ to $8n+1$ exclusive, with latent frame $0$ mapped to video frame $0$. This mapping provides a clean proxy for temporal editing in latent space at an 8-frame resolution.

To mitigate visual artifacts introduced by frame removal, we apply additional noise to latent frames adjacent to removed segments, allowing the diffusion process to smoothly regenerate motion. Selective re-rendering is implemented by setting the spatial mask $\mathbf{M=1}$ at arbitrary regions--such as the face, head or hands--so that only those regions can be noised and regenerated while the remainder of the video remains unchanged. Since editing alters the sequence length, temporal rotary positional embeddings (RoPE) are computed on the edited latent sequence. Finally, newly inserted frames are fully unmasked ($\mathbf{M=1}$), while existing frames retain face-region masking during lip-sync generation.

\begin{figure}[h]
    \centering
    \includegraphics[width=0.9\linewidth]{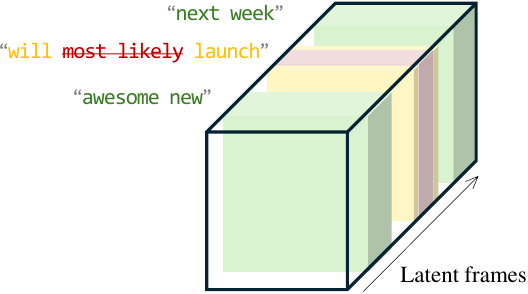}
    \caption{Timeline editing via manipulating latent frames.
    \textbf{Addition}: Insert \textcolor[HTML]{3B7D23}{new latents} with full noise. \textbf{Removal}: Discard \textcolor[HTML]{C00000}{corresponding latents} and noise \textcolor[HTML]{FFC000}{adjacent latents} to \textbf{re-render} a smooth transition. }
    \label{fig:edits}
\end{figure}

The overall latent-space editing process is illustrated in Figure~\ref{fig:edits}.

\subsection{Identity-Preserving Long Inference}
\begin{figure}[h]
    \centering
    \includegraphics[width=0.95\linewidth]{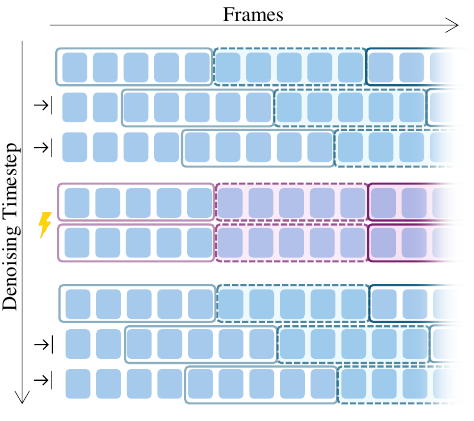}
    \caption{Long inference strategy: video latent frames are grouped into inference blocks and denoised iteratively. We apply a position shift to the blocks after each denoising step (blue inference blocks) to propagate context over longer windows throughout denoising. During medial timesteps (purple inference blocks) we disable the shift to benefit from TeaCache.}
    \label{fig:long_inference}
\end{figure}

\paragraph{Long Inference.}
\label{subsec:long_inference}
For long video generation, the sequence of latent frames is processed in blocks. Rather than fully denoising each block before proceeding to the next one (autoregressive long inference), we adapt the \textit{Time-aware position shift fusion} (TAPSF) long inference strategy proposed in Sonic~\cite{sonic,hunyuanvideoavatar}. We first encode the entire video into latent space and logically partition the full latent video into non-overlapping inference blocks. We choose a block width of $17$ latent frames ($136$ video frames).
For each timestep, we iteratively perform a \textbf{single} denoising step on each block of frames. For the next denoising timestep, the partition of frame latents into blocks is offset, such that the next denoising step will integrate context from adjacent frames, sharing longer-form context over many such denoising and offsetting steps (\cref{fig:long_inference}). The model then naturally bridges context between adjacent blocks throughout the entire denoising process, increasing inter-block stability. This strategy switches the order of looping between denoising steps and frame blocks compared to typical autoregressive-style inference. Further details can be found in~\cite{sonic}. 

At the video boundaries, frames that fall outside a block are handled by evaluating the overlapping regions twice: once from each neighboring block and averaging the resulting predicted velocities. RoPE are computed in the global frame coordinate space, ensuring consistent temporal positioning across shifts.

One disadvantage of TAPSF is its incompatibility with popular cache-based acceleration techniques such as~\cite{teacache,zhou2025easycache}. These methods accelerate inference by re-using the DiT blocks' outputs across previous timesteps if the inputs are ``similar'' enough. Similarity of the hidden states can be determined by, \eg a rescaled difference norm of the timestep-embedding-modulated inputs. With TAPSF, the previously computed block outputs correspond to a temporally \textit{shifted} set of features rather than a stationary representation of the same frames at a different noise level. 

Observing that lip synchronization, identity cues, and large-scale motion are primarily determined during early denoising timesteps, we adjust TAPSF to not shift blocks during the middle steps of denoising, and maintain a shift of $5$ latent frames during the early denoising steps (which are heavy on lip-sync, identity, and large motion) and late denoising steps (fine details). This modification allows us to apply adaptive caching during the middle $75\%$ of denoising steps, resulting in a speedup of approximately $1.6\times$ while preserving the benefits of TAPSF for long-range temporal coherence.

\paragraph{Identity Conditioning.}
\label{subsec:id_cond}
\begin{figure}[h]
    \centering
    \includegraphics[width=\linewidth]{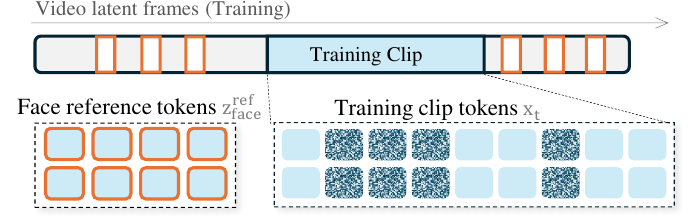}
    \caption{Identity conditioning: we train the DiT to use unnoised face tokens from outside the training clip to better preserve subject identity. These reference tokens are taken from a temporal neighborhood of the training clip and randomly added to the DiT's input sequence.}
    \label{fig:id_cond}
\end{figure}

Popular portrait animation methods~\cite{hunyuanportrait, fantasytalking, stableavatar, multitalk} address drift in subject identity by injecting facial features that capture the speaker’s visual characteristics, such as CLIP~\cite{CLIP}, DiNoV2~\cite{oquab2023dinov2}, or face embeddings~\cite{arcface} into dedicated cross-attention layers. For I2V models, the image prompt corresponding to the first frame of the video also implicitly serves as the identity reference. In this setup, the first frame conditions the model along two paths: through self-attention, as its clean tokens are present in the sequence of tokens entering the DiT and all the noisy tokens attend to it. In addition, features from the reference image can be injected through cross-attention.

In the V2V setting, a full reference \textit{video} of the subject is available during both training and inference. InfiniteTalk~\cite{infinitetalk} dynamically swaps the single reference frame for each inference block, with a frame from the video, to preserve appearance and coarse temporal progression in ``sparse video-to-video dubbing.'' However, we aim to leverage the subject’s identity and speaking style present throughout the entire video, rather than reducing conditioning to a single frame at a time. 

To this end, we fine-tune the self-attention mechanism to condition on reference frames. During training, we randomly sample 6 latent frames (corresponding to 64 video frames) from a temporal window of $\pm5\,$s around the target clip, encode them, and retain only tokens corresponding to the lower face region. These \textit{face reference tokens} $\mathbf{z}_{\text{face}}^{\text{ref}}$
are kept un-noised and concatenated to the video tokens along the sequence dimension.

OmniHuman-1~\cite{omnihuman_1} also conditions on a reference frame as concatenation to the token sequence, specifically by zeroing the temporal component of its 3D Rotary Positional Embedding (RoPE), effectively removing temporal ordering and motion information while still providing appearance cues.
While this design suffices for a single reference frame, it fails to extend to our reference-video setting, where multiple reference frames correspond to the same spatial region (e.g., the mouth) but capture different temporal states.
Zeroing their temporal embeddings would align all reference tokens on the same RoPE phase, leading to {aggregation bias}---the model tends to average or equally attend to all reference tokens, despite each representing visuals for distinct phonemes or lip positions.
To mitigate this, we assign unique sentinel temporal indices ($t=-1,-2,\ldots$) to reference tokens from different frames.
This preserves their distinct temporal identities while keeping them separable from the generated video’s temporal sequence. In inference, we sample face reference tokens from the block processed if available, and optionally from adjacent blocks.

\begin{figure}[h]
    \centering
    \includegraphics[width=\linewidth]{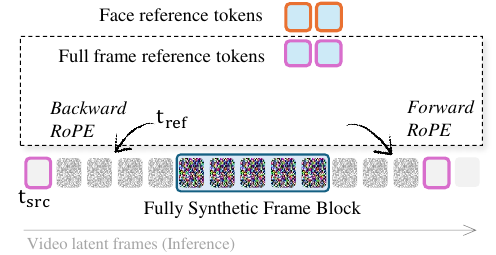}
    \caption{During inference of fully-synthetic blocks (i.e., blocks without V2V or first-frame condition), we prevent global appearance drift by adding full-frame reference tokens (the closest past and future latent frames) to the input sequence. These reference frames are not noised and their temporal indices are adjusted such that the temporal distance between the reference frames and the block is not greater than 3.}
    \label{fig:fw_bk_rope}
\end{figure}

For inference on fully synthetic blocks lacking clean video latents, as in long I2V generation or extended additions, we must also prevent drift in global frame appearance. To this end, We propose \textit{forward-backward RoPE conditioning} during inference, assigning full-frame reference tokens $\mathbf{z}_{\text{frame}}^{\text{ref}}$ from the last-seen and first-available future clean latent frames (see Fig.~\ref{fig:fw_bk_rope}). These tokens are assigned ``fake'' temporal indices $t_{\text{ref}}$ to provide appearance cues that are aligned enough in the temporal phase of RoPE without forcing exact replication of those frames. i.e $\text{RoPE}(\mathbf{z}_{\text{frame}}^{\text{ref}}, t_{\text{ref}})$ with
\begin{equation}
t_{\text{ref}} = 
\begin{cases} 
t_{\text{source}} & \text{if } \Delta t \leq 3 \\
t_{\text{block (end)}} + 3 & \text{if } \Delta t > 3 \text{ (forward)} \\
t_{\text{block (start)}} - 3 & \text{if } \Delta t > 3 \text{ (backward)}
\end{cases}
\end{equation}

where $\Delta t = |t_{\text{source}} - t_{\text{block}}|$ is the temporal distance between the source reference frame and the frame block boundary. Similar to the face reference tokens, the frame reference tokens $\mathbf{z}_{\text{frame}}^{\text{ref}}$ are also concatenated with the video tokens along the sequence dimension for inference.

We note that adapting the temporal embedding of frames as a method to address consistency is also proposed in concurrent work~\cite{lookahead, huang2025liveavatarstreamingrealtime}.

Combined with our long-inference approach, our model can output minutes-long videos without noticeable identity drift (see \href{https:\projectpage}{project page}).

\subsection{Training Loss}
Our final training loss becomes 
\begin{equation}
\label{eq:fulltrainingobjective}
\mathcal{L}_{\text{FM}} = \mathbb{E}_{t, \epsilon, \mathbf{x}_0, \mathbf{c}} [\| \textbf{M} \cdot v_\theta(\mathbf{z}_{\text{in}}, t, \mathbf{c}_{\text{audio}}, \mathbf{c}_{\text{text}}, \textbf{M}) - \mathbf{u}_t \|^2_2]
\end{equation}
where the target velocity $\mathbf{u}_t$ is masked to focus the learning signal on the mouth region:$$\mathbf{u}_t = \textbf{M} \odot (\epsilon - \mathbf{x}_0)$$ 
with the input tokens $\mathbf{z}_{\text{in}} = [\mathbf{z}_{\text{face}}^{\text{ref}}, \mathbf{z}_{\text{frame}}^{\text{ref}}, \mathbf{x}_t]$, containing the face reference tokens, frame reference tokens, and noisy video tokens, all concatenated along the sequence dimension.

\section{Experiments}
\label{sec:experiments}
\subsection{Training}
\label{subsec:training}

We base our model on the Lightricks \texttt{LTX-0.9.7} architecture and open-sourced weights~\cite{ltx,ltx_github}. 

\begin{figure}[h]
    \centering
    \includegraphics[width=1.\linewidth]{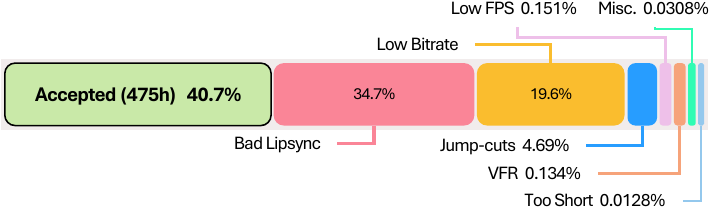}
    \caption{Data filtering results: a significant share of in-the-wild short-form videos exhibited low visual quality or bad lip-sync. We removed these from the dataset to achieve optimal training performance.}
    \label{fig:data-filter}
\end{figure}

\paragraph{Dataset.} We collect a total of 1,070 hours of talking-head footage. This includes 70 hours of proprietary, high-quality frontal recordings, along with 1,000 hours of shorter user-generated content gathered from YouTube, exhibiting substantial variability in appearance, identity, pose, background, gestural dynamics, and composition. The videos span a broad range of resolutions ($0.25$ to $2.0$~MP), aspect ratios $(2:1, 1.78:1, \dots, 1:1, \dots, 0.56:1, 0.5:1)$, and frame rates (24–60 fps). This enables support for diverse input formats during inference. Additionally, we filter the dataset for scene cuts, lip-sync confidence score ($\text{SyncC}\geq3$~\cite{wav2lip}), and temporal offset ($\leq40 ms$)~\cite{syncnet}, bit-rate ($\geq2000\text{Kbps}$), frame-rate ($\in [24-60]$), number of frames ($\geq121$) and corrupted videos, finally yielding 475 hours of video cut into 121-frame video clips. The data filtration process is depicted in Fig~\ref{fig:data-filter}. We generate video captions with CogVLM2~\cite{cogvlm2}.

\paragraph{Model Training.} Training proceeds in two stages. In stage 1, the new audio projection module and cross-attention layers are optimized (with all other weights frozen) for $20k$ steps, encouraging the model to learn lip synchronization without compromising the generative prior of the pre-trained DiT. In Stage 2, training transitions to a 128-rank low-rank adaptation (LoRA) of the full DiT for another $10k$ iterations, which enables identity conditioning, further improves lip synchronization, and is instrumental to the visual fidelity of the resulting videos.

We train our model to generate videos with any combination of input conditions and at a range of video frame rates and resolutions. To support varied input combinations, we randomly enable the video-to-video, first-frame, and audio conditions with probabilities $p_\mathrm{v2v}$, $p_\mathrm{ff}$ and $p_\mathrm{audio}$, respectively (see Table \ref{tab:training_parameters}). Note that audio conditioning is always enabled in the first training stage since the unconditional case reduces to the base DiT model. We enable identity reference conditioning with probability $p_\mathrm{id}$ only during the second stage. To support multiple video resolutions, we randomly resize each video clip to a resolution between 0.25 and 2.0 megapixels, then trim the number of frames to maintain an equal number of video tokens per batch. The original frame rate of each video clip is encoded in the 3D RoPE, which is computed separately for each batch element.

\begin{table}[t]
\centering
\small
\caption{Hyperparameters across training stages.}
\label{tab:training_parameters}
\begin{tabular}{lcc}
\toprule
\textbf{\small Parameter} & \textbf{\small Stage 1: Audio-Only} & \textbf{\small Stage 2: LoRA} \\
\midrule
Steps         & $20k$ & $10k$ \\
Base LR: $lr_{\mathrm{adam}}$         & $1\mathrm{e}{-5}$ & $1\mathrm{e}{-4}$ \\
Audio Cond:       $p_{\mathrm{audio}}$        & 1.0 & 0.9 \\
FF Cond: $p_{\mathrm{ff}}$  & 0.9 & 0.9 \\
V2V Cond:         $p_{\mathrm{v2v}}$          & 0.9 & 0.9 \\
ID Cond:          $p_{\mathrm{id}}$     & 0.0 & 0.5 \\
\bottomrule
\end{tabular}
\end{table}

We use the Muon Optimizer~\cite{muon_opt} for matrix-shaped parameters and AdamW~\cite{adamw} for the remaining parameters which improved lip-sync and lowered our flow matching loss substantially in the first stage. We use learning rate $lr_\text{adam}$ for AdamW and $lr_\text{muon}=100\times lr_\text{adam}$ for Muon and increase $lr_\text{adam}$ and $lr_\text{muon}$ by $10\times$ in the second stage. We train for 42 hours on $8\times$ H100 GPUs with a batch size of 4 for each GPU.

To improve training efficiency and stability, we incorporate two complementary techniques: (1) immiscible diffusion with KNN-based noise selection ($k=4$)~\cite{li2025immiscible}, which reduces diffusion trajectory mixing and accelerates convergence, and (2) contrastive flow matching~\cite{stoica2025contrastive}, which enforces uniqueness across conditional flows to enhance audio-visual correspondence and identity preservation. These techniques work synergistically - immiscible diffusion reduces trajectory miscibility while contrastive flow matching explicitly maximizes dissimilarities between flow from different audio conditions, helping the model better distinguish between different audio features and their corresponding visual representations across the entire denoised region.

\subsection{Evaluation}
We evaluate our model’s video generation performance in terms of lip synchronization and visual fidelity across both I2V and V2V settings. Extensive qualitative comparisons are provided on our \href{https://\projectpage}{project page}, which we recommend viewing to assess results in their native video format. We therefore focus on quantitative evaluation below.

\paragraph{Video-to-Video.}
To evaluate a re-render of an existing video with a new audio track, we first consider a controlled reconstruction (self-reenactment) setting where ground truth video is available. This setup measures the model’s ability to preserve visual fidelity and speaker identity, including mouth shape, facial details, and temporal dynamics. We evaluate on a subset of 100 videos\footnote{For InfiniteTalk we reduce to 20 videos due to long runtimes.} from the TalkVid dataset~\cite{talkvid}, and compare against several state-of-the-art V2V lip-sync systems. These baselines include both open-source research models and widely deployed commercial solutions.

In addition to self-reenactment, we evaluate a re-render with novel audio by pairing each source video with audio from a different TalkVid video, measuring the model’s ability to preserve visual fidelity while accurately synchronizing unseen speech.

We report standard image and video quality metrics FID~\cite{heusel2018ganstrainedtimescaleupdate} and FVD~\cite{unterthiner2019accurategenerativemodelsvideo} as well as identity preservation (CSIM~\cite{ghazouali2024csimcopulabasedsimilarityindex}) and lip-sync accuracy (Sync-C and Sync-D~\cite{syncnet}). Results are summarized in Table~\ref{tab:v2v_results}.

\begin{table*}[t]
    \centering
    \small
    \begin{minipage}{0.9\textwidth} 
        \centering
        \caption{Quantitative results on \textbf{Video-to-Video} lip-syncing evaluated on the \textbf{TalkVid}~\cite{talkvid} dataset. We compare methods on \textbf{Novel Audio} (audio from a different video) and \textbf{Self-Reenactment} (audio from the source video). Metrics include \textbf{FID} and \textbf{FVD} (image/video fidelity $\downarrow$), \textbf{CSIM} (identity preservation $\uparrow$), and \textbf{Sync-C/D} (lip-sync confidence $\uparrow$ and distance $\downarrow$). \textbf{Pose Preservation} indicates if the method retains the original head pose. We highlight \colorbox{bestc}{\textbf{best}} and \colorbox{secondc}{\textbf{second best}} performance.}
        \label{tab:v2v_results}
        \setlength{\tabcolsep}{5pt}
        \begin{tabular*}{\textwidth}{@{\extracolsep{\fill}}l c c c c c c}
            \toprule
            \textbf{Method} & \textbf{FID $\downarrow$} & \textbf{FVD $\downarrow$} & \textbf{CSIM $\uparrow$} & \textbf{Sync-C $\uparrow$} & \textbf{Sync-D $\downarrow$} & \textbf{Pose Preservation} \\
            \midrule
            
            \multicolumn{7}{l}{\textit{\textbf{Video-to-Video (Self-Reenactment)}}} \\
            \midrule
            
            \multicolumn{7}{l}{\hspace{3mm}\textit{\textbf{Open Source}}} \\
            \addlinespace[3pt]
            \hspace{6mm} LatentSync~\cite{latentsync}       & 46.22 & 148.19 & 0.88 & 7.11 & 1.559 & \checkmark \\
            \hspace{6mm} InfiniteTalk~\cite{infinitetalk}   & 53.12 & 304.04 & 0.86 & 7.31 & 1.538 & \ding{55} \\ 
            \hspace{6mm} MuseTalk~\cite{musetalk}           & 47.78 & 134.46 & 0.82 & 5.43 & 1.954 & \checkmark \\
            
            \addlinespace[5pt]
            
            \multicolumn{7}{l}{\hspace{3mm}\textit{\textbf{Commercial}}} \\
            \addlinespace[3pt]
            \hspace{6mm} Pixverse V5 Lipsync                & 45.94 & \second{116.54} & 0.89 & 5.67 & 1.880 & \checkmark \\
            \hspace{6mm} Sync.so React 1                    & 48.07 & 161.04 & 0.86 & 6.59 & 1.604 & \checkmark \\
            \hspace{6mm} Sync.so V2 Pro                     & \best{35.51} & 109.88 & \second{0.91} & \second{7.43} & \second{1.547} & \checkmark \\
            \hspace{6mm} Veed Lipsync                       & 53.02 & 171.94 & 0.86 & 6.74 & 1.637 & \checkmark \\
            \hspace{6mm} Creatify Lipsync                   & 64.88 & 341.47 & 0.69 & 7.04 & 1.585 & \checkmark \\
            
            \midrule 
            
            \hspace{6mm} \textbf{Ours}                      & \second{37.10} & \best{109.04} & \best{0.92} & \best{7.50} & \best{1.480} & \checkmark \\
            
            \midrule
            \midrule
            
            \multicolumn{7}{l}{\textit{\textbf{Video-to-Video (Novel Audio)}}} \\
            \addlinespace[3pt]
            \hspace{6mm} LatentSync~\cite{latentsync}       & 56.84 & 157.73 & \second{0.88} & \second{6.90} & 1.857 & \checkmark \\
            \hspace{6mm} InfiniteTalk~\cite{infinitetalk}   & \second{42.93} & 286.03 & \second{0.88} & 6.89 & \second{1.568} & \ding{55} \\
            \hspace{6mm} MuseTalk~\cite{musetalk}           & 49.59 & \second{140.49} & 0.82 & 5.04 & 1.922 & \checkmark \\
            
            \midrule 
            
            \hspace{6mm} \textbf{Ours}                      & \best{41.25} & \best{104.18} & \best{0.89} & \best{7.36} & \best{1.502} & \checkmark \\
            
            \bottomrule
        \end{tabular*}
        \vspace{0.8em}
        \footnotesize
    \end{minipage}
\end{table*}

\begin{table*}[t]
    \small
    \centering
    \begin{minipage}{1.0\textwidth} 
        \centering
        \caption{Quantitative results on \textbf{Image-to-Video} lip-syncing evaluated on \textbf{TalkVid}~\cite{talkvid} and \textbf{VBench}~\cite{vbench}. We report standard lip-sync metrics (definitions follow Table~\ref{tab:v2v_results}) and general video quality metrics: \textbf{Subj./Back.} (subject/background consistency $\uparrow$), \textbf{Aesth.} (aesthetic quality $\uparrow$), and \textbf{Motion} (smoothness $\uparrow$). We highlight \colorbox{bestc}{\textbf{best}} and \colorbox{secondc}{\textbf{second best}} performance.}
        \label{tab:i2v_results}
        
        \setlength{\tabcolsep}{3.5pt} 
        
        \begin{tabular*}{\textwidth}
        {@{\extracolsep{\fill}}l ccccc cccc}
            \toprule
            & \multicolumn{5}{c}{\textbf{\textit{Image-to-Video}: TalkVid}~\cite{talkvid}} & \multicolumn{4}{c}{\textbf{\textit{Image-to-Video}: VBench}~\cite{vbench}} \\
            \cmidrule(lr){2-6} \cmidrule(lr){7-10}
            
            \textbf{Method} & \textbf{FID} $\downarrow$ & \textbf{FVD} $\downarrow$ & \textbf{CSIM} $\uparrow$ & \textbf{Sync-C} $\uparrow$ & \textbf{Sync-D} $\downarrow$ & \textbf{Subj.} $\uparrow$ & \textbf{Back.} $\uparrow$ & \textbf{Aesth.} $\uparrow$ & \textbf{Motion} $\uparrow$ \\
            \midrule
            
            \multicolumn{10}{l}{\textit{\textbf{Open Source}}} \\
            \addlinespace[3pt]
            \hspace{3mm} Hallo3~\cite{hallo3}       
            & 68.11 & 524.05 & 0.79 & 6.25 & 1.679 
            & 0.939 & 0.941 & 0.306 & 0.984 \\
            
            \hspace{3mm} InfiniteTalk~\cite{infinitetalk} 
            & \second{55.47} & \best{285.75} & \best{0.86} & 6.75 & 1.642 
            & \second{0.975} & 0.950 & 0.445 & \best{0.992} \\
            
            \hspace{3mm} Sonic~\cite{sonic}         
            & 63.66 & 413.54 & 0.85 & 6.98 & 1.629 
            & 0.969 & 0.936 & 0.400 & 0.985 \\
            
            \hspace{3mm} StableAvatar~\cite{stableavatar} 
            & 65.56 & 459.33 & 0.79 & 6.39 & 1.683 
            & 0.958 & 0.936 & 0.318 & 0.990 \\
            
            \midrule
            
            \multicolumn{10}{l}{\textit{\textbf{Commercial}}} \\
            \addlinespace[3pt]
            \hspace{3mm} Creatify Aurora 
            & 82.83 & 332.77 & 0.74 & 6.53 & \second{1.606} 
            & 0.972 & 0.937 & 0.419 & 0.991 \\
            
            \hspace{3mm} Veed Fabric     
            & 68.59 & 439.78 & 0.82 & 6.03 & 1.898 
            & 0.973 & 0.948 & \best{0.495} & 0.991 \\
            
            \hspace{3mm} Kling Pro V2    
            & 72.32 & 547.28 & 0.82 & 6.12 & 1.843 
            & \best{0.977} & \best{0.953} & 0.472 & 0.990 \\
            
            \hspace{3mm} OmniHuman V1.5  
            & 76.68 & 363.13 & 0.78 & 5.27 & 2.307 
            & 0.960 & 0.939 & 0.451 & 0.989 \\
            
            \hspace{3mm} Fal/HunyuanVideo-Avatar~\cite{hunyuanvideoavatar}
            & 57.05 & 383.17 & 0.81 & 5.04 & 2.372 
            & 0.968 & 0.935 & 0.427 & 0.986 \\
            
            \hspace{3mm} Fal/MultiTalk~\cite{multitalk}
            & 65.69 & 312.69 & 0.83 & 6.50 & 1.789 
            & 0.956 & 0.929 & 0.435 & 0.989 \\
            
            \hspace{3mm} Fal/StableAvatar~\cite{stableavatar}
            & 68.52 & 593.48 & 0.77 & 4.61 & 2.939 
            & 0.974 & 0.951 & 0.459 & 0.985 \\
            
            \midrule
            
            \hspace{3mm} \textbf{Ours} 
            & \best{55.38} & \second{312.46} & \best{0.86} & \best{7.21} & \best{1.516} 
            & 0.973 & \best{0.953} & \second{0.486} & \best{0.992} \\
            
            \bottomrule
        \end{tabular*}
    \end{minipage}
\end{table*}

\paragraph{Image-to-Video.}
To evaluate performance in the I2V setting, we condition on the first frame of each TalkVid video and the first four seconds of the corresponding audio track. We compare against a range of recent I2V talking-head generation methods, including both open-source research models and commercial systems.

In addition to lip-sync and fidelity metrics, we report VBench~\cite{vbench} evaluation scores for Subject Consistency, Background Consistency, Aesthetic Quality, and Motion Smoothness on a subset of 30 videos using a one-minute audio track.

\subsection{Performance Optimizations}
With all of the following optimizations enabled, our model renders a 10-second 1080p video in 225 seconds on a single H100 GPU. In comparison, InfiniteTalk, while comparable in image fidelity, requires approximately 10,000 seconds to generate the same video. These results demonstrate that our system achieves practical inference speeds suitable for real-world, long-form video editing workflows.
\paragraph{VAE Tiling and Latent Frame Blocking.}
For memory optimization, we implement temporal tiling with an overlap of 16 video frames for the VAE encoding and decoding. 
For denoising, we choose a block size of 17 latent frames (corresponding to 136 video frames), which balances memory efficiency with sufficient temporal context for stable generation. This block structure is consistent with the shifting strategy described in Section~\ref{subsec:long_inference} and enables scalable inference on long sequences.

\paragraph{Quantization.}
To optimize inference speed without compromising quality improvements from the fine-tuning stage, we employ a hybrid FP8 quantization strategy. Specifically, we quantize the large pre-trained base weights in the FFN and Attention layers, while preserving LoRA adapters and architectural bottlenecks \eg final projection layers and embeddings in BF16 precision. This selective quantization yields a $\times2.5$ speedup while maintaining output quality.
\paragraph{\textbf{Hybrid Sequence Parallelism.}}
To eliminate computational bottlenecks in the attention layers during high-resolution generation, we adopt Hybrid Sequence Parallelism by combining Ulysses and Ring Attention~\cite{fang2024uspunifiedsequenceparallelism}. This approach distributes the attention computation across multiple GPUs, improving throughput without increasing memory pressure. When deployed on an 8$\times$H100 GPU node, this strategy provides an aggregate $\times8$ speedup within the denoising loop.

\subsection{Ablation Study}
\begin{figure}[h]
    \centering
    \includegraphics[trim={0 0.5cm 0 0},clip,width=1.0\linewidth]{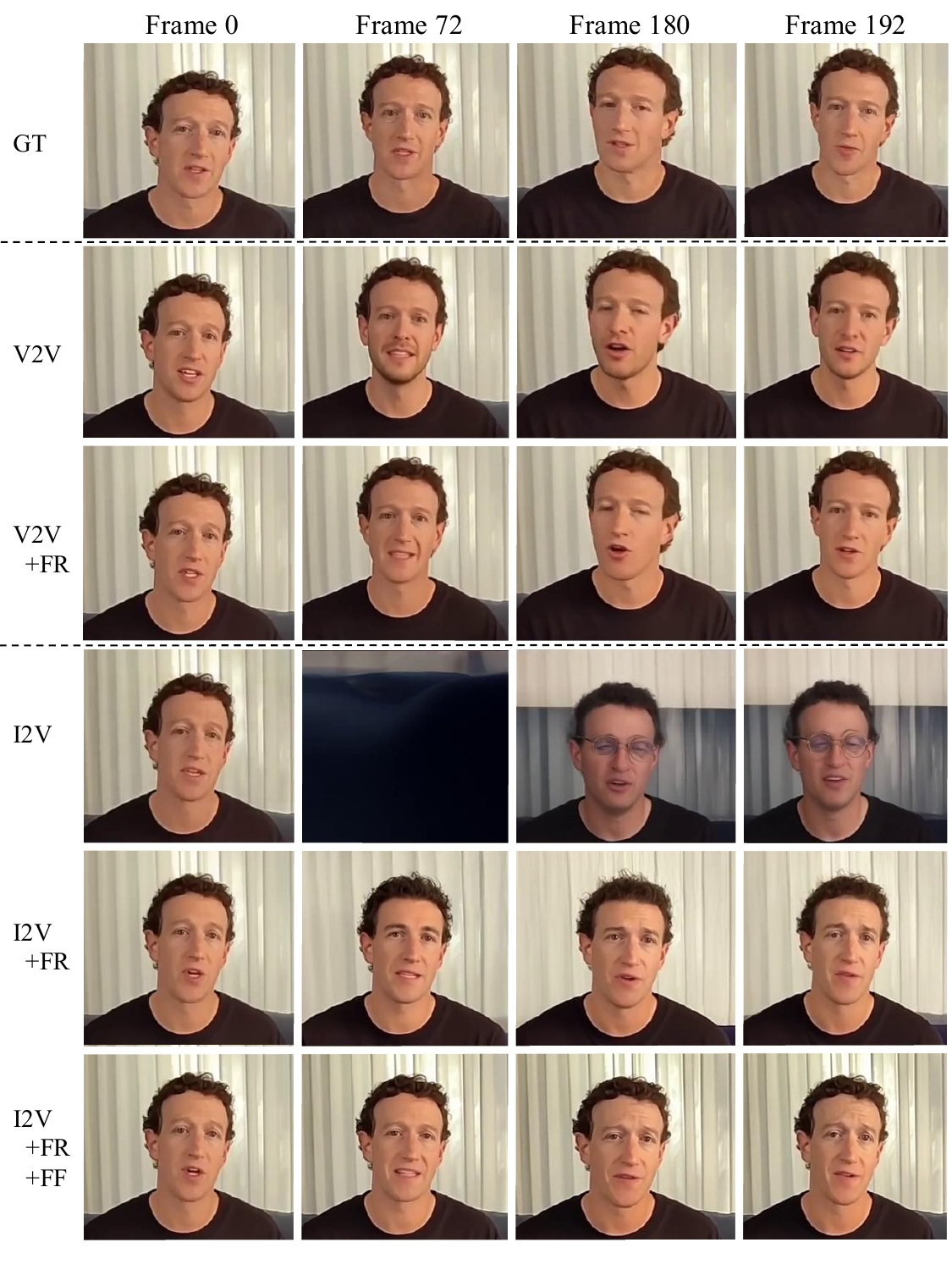}
    \caption{Ablation of the reference frame conditioning.
    From left to right, we show 4 representative frames from an 8-second clip rendered with different reference conditioning variants. From top to bottom: groundtruth (GT), pure video-to-video (V2V), video-to-video with face reference tokens (V2V+FR), pure image-to-video (I2V), image-to-video with face reference tokens (I2V+FR),  image-to-video with face and full-frame reference tokens (I2V+FR+FF).}
    \label{fig:ablation-idref}
\end{figure}

\paragraph{What happens if Identity conditioning is dropped?} 
Without reference tokens, the subject's appearance drifts away rather quickly from its original state since the DiT can access the first-frame condition only in the first and last block (due to circular padding~\cite{sonic}). All intermediate inference blocks have no direct access to clean lower face tokens as a reference.
Our TAPSF long inference strategy softens the impact of drift as appearance information is shared between neighboring blocks over the course of the denoising process, which finally results in a smooth appearance drift that is most obvious in the middle of the video (\cref{fig:ablation-idref}, V2V-Frame72).
In the I2V case (latent frames are fully noised), this drift becomes more severe, resulting in a complete change of appearance and scene (\cref{fig:ablation-idref}, I2V-Frame72).
\cref{fig:ablation-idref} visualizes the benefit of using face (FR) and full-frame (FF) reference tokens during inference.
Pure V2V exhibits a clear identity drift (beard growth) after just 72 frames, whereas V2V+FR maintains the original identity well throughout the video (\cref{fig:ablation-idref}). 
While in the video-to-video case, only the person's face and small background details are affected, fully synthetic sections (e.g., long additions or completely re-rendered sections) suffer from obvious discontinuities even at the scene level (\cref{fig:ablation-idref}, I2V-Frame72). I2V+FR maintains the appearance of the lower face, but shows a mild scene drift (curtains, hair, eyebrows) since the face reference tokens contain only the lower face and a small portion of the background. I2V+FR+FF maintains full scene and identity consistency while being able to fully re-render the captured performance with novel head poses and facial expressions (eyebrows, forehead), see \cref{fig:ablation-idref} (bottom row).

\begin{figure}[h]
    \centering
    \includegraphics[width=1.0\linewidth]{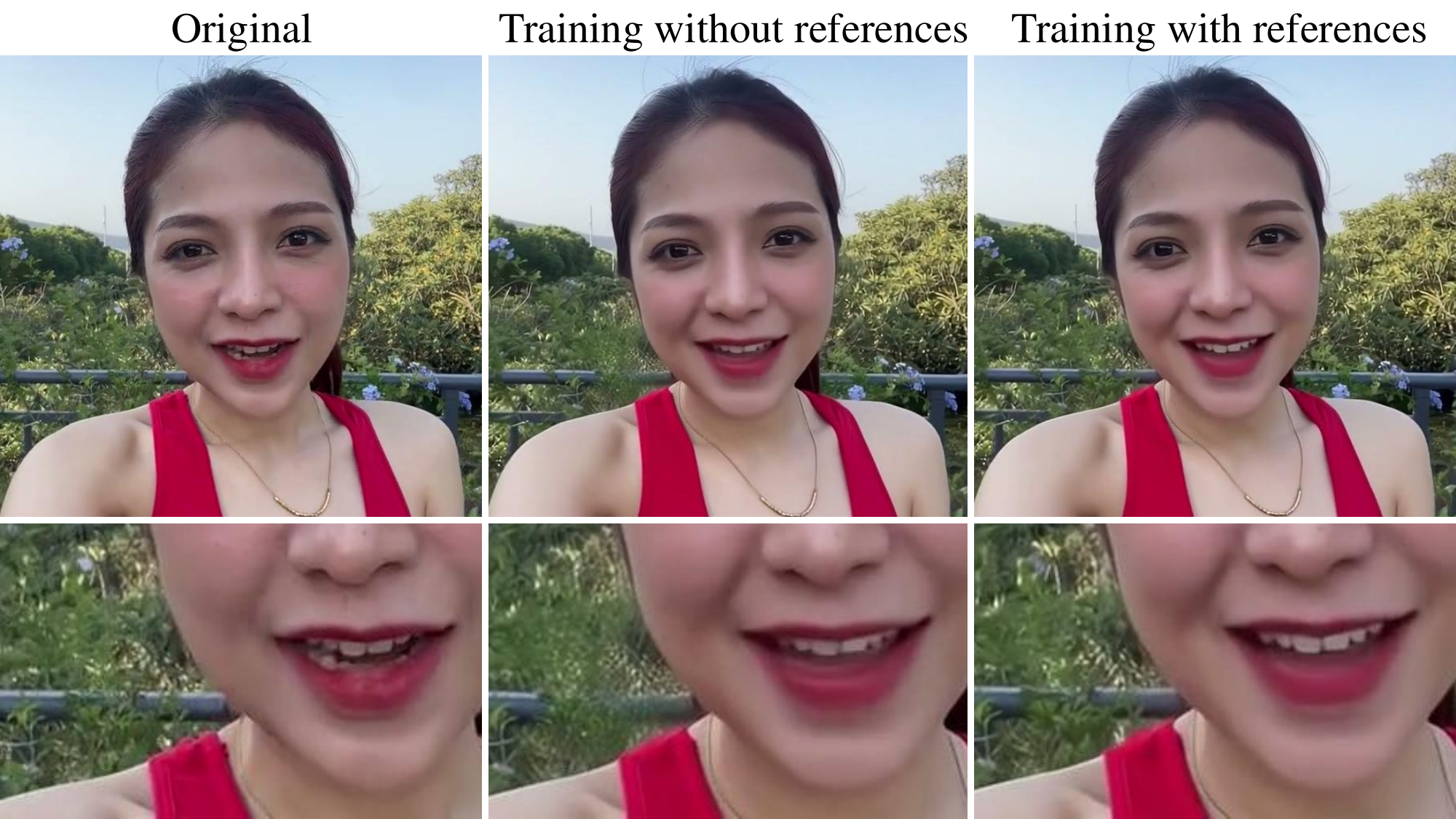}
    \caption{The difference in render quality with and without exposing the DiT to reference tokens during training. Left: ground truth, middle: no reference tokens during training, right: with reference tokens during training. Without training for reference tokens at sentinel timesteps, render artifacts appear, particularly with complex and dynamic backgrounds.}
    \label{fig:ablation-idref-training}
\end{figure}

\paragraph{Training with/without Identity Reference Condition.}
While the DiT is capable of utilizing face reference tokens without having been exposed to them during training, incorporating reference tokens during training with a probability of 50\% leads to improved rendering quality, particularly in complex scenes with highly structured and dynamic backgrounds (\cref{fig:ablation-idref-training}).

\section{Conclusion}
\label{sec:conConclusionclusion}
In summary, we present \paper, a diffusion-based framework for audio-driven talking head synthesis that extends a general-purpose video diffusion model with audio-driven V2V editing capabilities. Through a two-stage training scheme and a windowed audio conditioning strategy, our approach enables precise lip synchronization while preserving visual fidelity to the original video content. We further introduce Forward-Backward RoPE Conditioning to maintain stable identity and appearance over extended durations. By operating directly in latent space, \paper{} enables transcript-based modification of videos, including addition, removal and retiming, offering a practical step toward using generative video models as tools for professional post-production, alongside their role in end-to-end synthesis.
\subsection{Ethical Considerations}

\paper's capacity for high-fidelity synthesis and the granular manipulation of existing footage necessitates careful consideration of potential misuse, particularly in the context of visual forgeries and the dissemination of misinformation. We emphasize that responsibility for the ethical use of these techniques is shared across the research community, including those who deploy or build upon methods introduced in this work. We therefore recommend a multi-layered approach to responsible deployment: (1) Establishing legal barriers such as explicit declarations of content ownership, and (2) implementing technical safeguards including celebrity detection, identity verification and robust digital watermarking. Furthermore, the authors strongly advocate for continued research into content provenance and synthetic media detection to mitigate risks associated with unauthorized generation and to ensure the ethical evolution of generative video tools.

\subsection{Acknowledgements}
The authors would like to thank the Lightricks LTX-Video team for open-sourcing their model weights, and specifically Ofir Bibi, Yoav HaCohen, and Nisan Chirput for their technical insights during the development of this work.

{
    \small
    \bibliographystyle{ieeenat_fullname}
    \bibliography{main}
}


\end{document}